\documentclass{article}

\usepackage{wrapfig}
\usepackage{tabularx}
\usepackage[table]{xcolor}
\usepackage{placeins}
\usepackage{float}
\usepackage{afterpage}
\usepackage{amsmath}
\usepackage{microtype}
\usepackage{graphicx}
\usepackage{subfigure}
\usepackage{booktabs}
\usepackage{hyperref}

\definecolor{lblue}{rgb}{0.610, 0.762, 0.812}
\definecolor{lightblue}{rgb}{0.776, 0.894, 0.945}
\usepackage[accepted]{icml2024}

\icmltitlerunning{OnionEval: An Unified Evaluation of Fact-conflicting Hallucination for Small-Large Language Models}

\begin{document}

\twocolumn[
\icmltitle{OnionEval: An Unified Evaluation of Fact-conflicting Hallucination for Small-Large Language Models
}



\icmlsetsymbol{equal}{*}

\begin{icmlauthorlist}
\icmlauthor{Chongren Sun}{mcgill}
\icmlauthor{Yuran Li}{mcgill}
\icmlauthor{Di Wu}{mcgill}
\icmlauthor{Benoit Boulet}{mcgill}
\end{icmlauthorlist}

\icmlaffiliation{mcgill}{Intelligent Automation Lab, McGill University}

\icmlcorrespondingauthor{Chongren Sun}{chongren.sun@mail.mcgill.ca}

\icmlkeywords{Machine Learning, ICML}

\vskip 0.3in
]



\printAffiliationsAndNotice{\icmlEqualContribution} 

\begin{abstract}
Large Language Models (LLMs) are highly capable but require substantial computational resources for both training and inference. Within the LLM family, smaller models (under 10 billion parameters) also demonstrate strong performance across various tasks. However, these smaller models still share the same limitations as their larger counterparts, including the issue of hallucination. Despite numerous benchmarks developed to evaluate hallucination in LLMs, few have focused specifically on small LLMs (SLLMs). Additionally, SLLMs exhibit significantly different performance across various benchmarks. In this paper, we introduce OnionEval, a multi-layer structured framework with a specific metric: context-influence score (\textit{CI}), designed to effectively assess the fact-conflicting hallucination tendencies of small LLMs across different contextual levels. Our experimental results highlight a key feature of SLLMs: they perform well on factual analysis but struggle with context reasoning. Our further investigation demonstrates simple Chain-of-thought strategy can greatly mitigate this limitation, enhancing the practical applicability of SLLMs in real-world scenarios.\footnote{Data and code are available at \url{https://github.com/sunchongren/OnionEval}}

\end{abstract}

\section{Introdcution}
\label{submission}
\begin{figure}[t]
\vskip 0.2in
\begin{center}
\centerline{\includegraphics[width=\columnwidth]{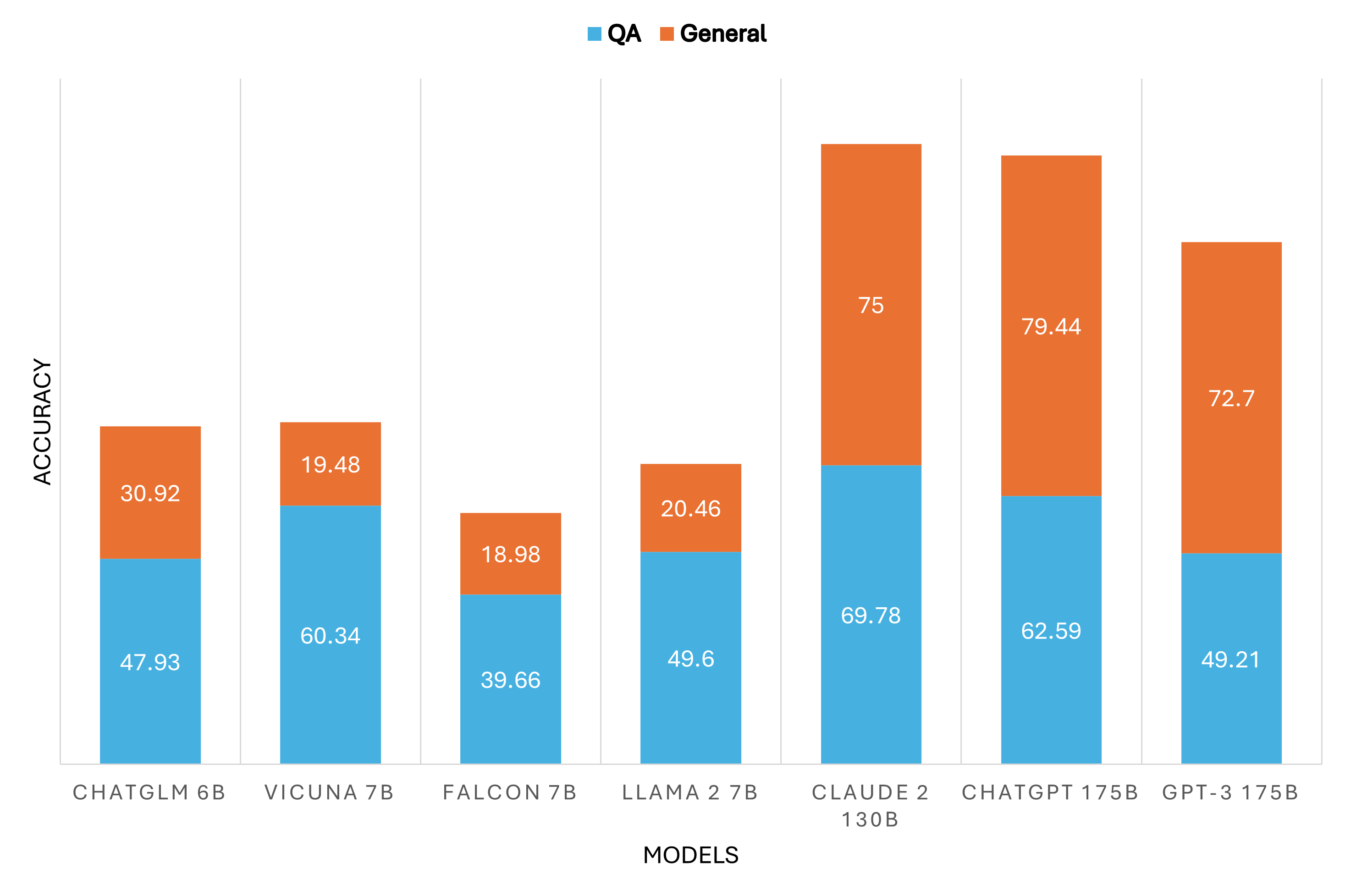}}
\caption{Halueval benchmark, SLLMs (in red rectangle) have similar performance on QA dataset, but largely lower performance on General dataset. Larger models' evaluation are more stable. This bring one question by us: why SLLMs perform vastly different on different benchmarks?}
\label{icml-historical}
\end{center}
\vskip -0.2in
\end{figure}
With advancements in transformer architectures, neural networks have scaled significantly beyond earlier models, ushering in a new era of Natural Language Generation (NLG) led by Large Language Models (LLM). LLMs, characterized by their high parameter counts, demonstrate remarkable capacity for understanding and generating complex, nuanced responses across a wide range of topics \cite{hadi2023survey,scr}. However, these models require substantial computational resources for both training and inference, limiting their accessibility and widespread deployment.

As a subset of LLMs, Small LLMs (SLLM), typically with parameter sizes ranging from 1 billion to 10 billion, offer a reduced computational footprint. With lower memory and power consumption, SLLMs are more suitable for deployment on standard hardware, including edge devices and mobile applications. Despite these advantages, SLLMs inherit several challenges common to general LLMs, including the persistent issue of model hallucination. Hallucination refers to the phenomenon where the model produces outputs that are factually inaccurate, logically inconsistent, or fabricated, despite their appearance of coherence and confidence. This behavior stems from the model's reliance on statistical correlations learned during training rather than an intrinsic understanding of the content.

\begin{table*}[t]
    \centering
    \setlength{\tabcolsep}{10pt} 
    \renewcommand{\arraystretch}{1.2} 
    \label{tab:comparison}
    \begin{tabular}{lccc}
        \toprule
        \textbf{Datasets} & \textbf{Methodology} & \textbf{Objective} & \textbf{Target Models}  \\
        \midrule
        FACTOID \cite{factoid} & Automatic Evaluation & Segment Factuality & General \\
        FACTOR \cite{factor} & Automatic Evaluation & Scalable fact evaluation & General  \\
        FactCHD \cite{factchd}&  Automatic Evaluation   &  Factuality Pattern & General \\
        HalluQA \cite{haluqa} & Inference Classifier & Chinese Hallucination & Chinese LLM \\
        HaluEval \cite{li2023halueval} & Inference Classifier & Hallucination & General \\
        FactScore \cite{min2023factscore} & Evidence Retrieval & Atomic Fact Percentage & General \\
        \midrule 
        OnionEval (Ours) & Layered Evaluation &  In-context Hallucination  & SLLM \\
        \bottomrule
    \end{tabular}
    \caption{Comparison with other fact-conflicting hallucination datasets. OnionEval is the only dataset that focuses on Small LLMs, with a context-level evaluation framework. Methodologies from survey \cite{halu_survey}.}
\end{table*}

While extensive research has been conducted on evaluating and mitigating hallucinations in LLMs \cite{tonmoy2024comprehensive, luo2024hallucination}, there are no benchmarks specifically designed for SLLMs, especially their vastly different performance on different tasks. For example, the Halueval benchmark \cite{li2023halueval} evaluates models of various sizes and shows that SLLMs perform almost as well as larger models on QA datasets. However, they exhibit significantly poorer performance on General datasets (Figure~\ref{submission}). Similarly, results from FactCHD \cite{factchd} indicate that smaller model achieve comparable performance in comparison dataset, but weak performance on set-operation dataset. This discrepancy raises a critical question: why do SLLMs perform well in some benchmarks but poorly in others?

We hypothesize two reasons for this inconsistency. Firstly, the near performance of SLLMs compared to their larger counterparts on certain benchmarks (e.g., the HaluEval QA dataset) demonstrates their capability for knowledge compression within parameters. Secondly, the significantly lower performance of SLLMs on datasets requiring contextual reasoning highlights their poor ability to understand context, leading to the generation of hallucinated answers.

To tackle these challenges and drive progress in understanding hallucinations in SLLMs, we introduce OnionEval: a benchmark and framework specifically designed to evaluate fact-conflicting hallucinations across nested context layers. OnionEval provides standardized evaluation criteria, emphasizing both atomic fact evaluation and context-based hallucination assessment. Unlike previous benchmarks, OnionEval is uniquely suited to capture and analyze the influence of context on fact-conflicting hallucinations. A key component of this framework is the proposed Context Influence Score (\textit{CI}), a novel metric that quantifies the extent to which contextual information impacts factual hallucination behavior. To the best of our knowledge, OnionEval and CI are the first framework explicitly developed to address this critical aspect of hallucination evaluation.

We evaluated four main model families, Llama\footnote{https://ai.meta.com/blog/meta-llama-3/} (developed by Meta), Qwen\footnote{https://qwenlm.github.io/} (developed by Alibaba), Gemma\footnote{https://ai.google.dev/gemma} (developed by Google), spanning parameter sizes from 405 million to 3 billion. Our analysis reveals that while SLLMs perform comparably to larger models when detecting hallucinations in isolated atomic facts. However, they struggle significantly with context comprehension. Specifically, SLLMs demonstrate competitive performance for standalone atomic facts but exhibit a marked decline in accuracy when these facts are embedded within complex contexts. Furthermore, CI scores highlight the limited ability of SLLMs to handle hallucinations influenced by context, underscoring the need for continued research and targeted evaluation frameworks for these models.

We further investigated various hallucination mitigation methodologies for SLLMs, including Chain-of-thought, Retrieval Augmented Generation, and Few-shot prompting. Our findings reveal that, similar to LLMs, certain strategies (e.g., chain-of-thought) can greatly enhance their performance on our benchmark with layered context. This illustrates the potential of SLLMs to achieve high generation accuracy with additional processing while requiring relatively low computational power.


\section{Related Work}

\subsection{Small Large Language Model}
Unlike traditional defined small language models (SLMs) \cite{slm, slm2}, which typically have 100M-5B parameters and are used in traditional Natural Language Generation (NLG) task. small Large Language Models (SLLMs) are defined as models with significantly more parameters. However, SLLMs have fewer parameters compared to larger models such as GPT-4 or LLaMA-405B. Generally, SLLMs fall within the range of 3 billion to 10 billion parameters, offering a balanced trade-off between model size and computational efficiency.

\subsection{Hallucination}

\begin{figure*}[t]
    \centering
    \includegraphics[width=\textwidth]{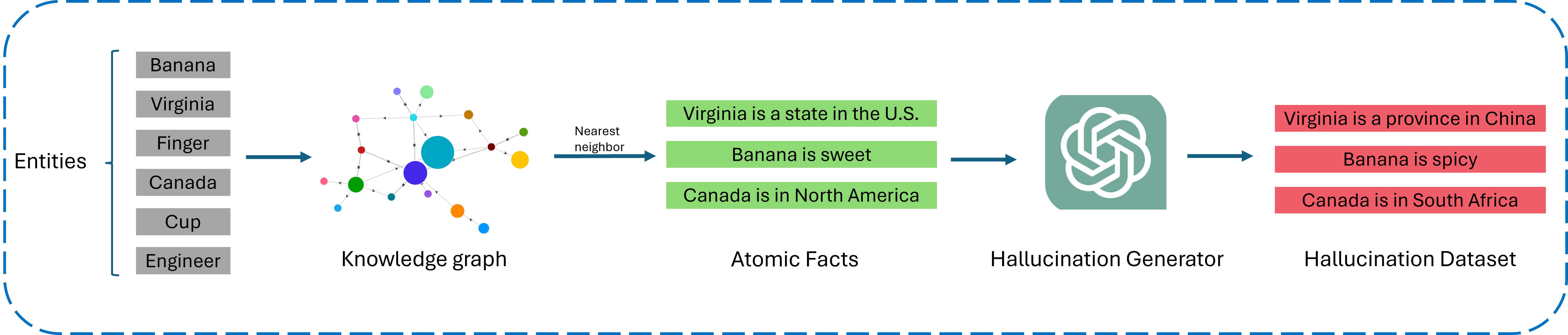}
    \caption{The construction pipeline of OnionEval. From left to right, they are: entity sampling, atmoic fact extraction, and hallucination generation}
    \label{fig:fixed-image}
\end{figure*}

Hallucinations in Large Language Models (LLMs) refer to instances where the model generates responses that diverge from the input, contradict earlier statements, or introduce incorrect information \cite{halu_survey}. Despite their impressive capabilities, LLMs sometimes produce outputs that appear coherent but are factually wrong or misleading. Recent research has identified three primary categories of hallucinations\cite{zhang2023siren}:

\textbf{Input-Conflicting Hallucination}: This type occurs when the generated response does not align with the user’s input or prompt, often due to task misinterpretation \cite{zhang2023siren}. For example, the model may generate unrelated content instead of a summary, indicating a failure to understand the specific request. Such errors are frequent in tasks like summarization and translation, where fidelity to the input is essential.

\textbf{Context-Conflicting Hallucination}: This category involves inconsistencies within the model’s own outputs, typically emerging in longer interactions where the model loses track of prior context \cite{zhang2023siren}. It results in contradictions to previous responses and undermines the coherence of dialogue-based tasks.

\textbf{Fact-Conflicting Hallucination}: LLM generates content that conflicts with widely accepted facts, leading to misinformation. Common causes of this type of hallucination are knowledge gap or instruction fine-tuning knowledge gap \cite{halu_survey}.This type of hallucination is particularly problematic in factual domains, such as question-answering systems, where the integrity of information is crucial.

\subsection{Hallucination benchmarks}

Recent research has introduced a range of benchmarks specifically designed to evaluate hallucination in large language models (LLMs). General hallucination benchmarks has two types: generation and discrimination. Generation benchmarks, like Truthfulqa \cite{lin2021truthfulqa}, often use coherence and fluency, with the help of human evaluator or LLM judge, as its metrics. For discrimination benchmarks, for example Halueval\cite{li2023halueval}, FACTOR\cite{factor}, are used to let llm judge a given input with a certain number of possible answers.

\subsection{Context-influenced hallucination}

With extensive pre-training data and deep, complex networks, Large Language Models (LLMs) have the capability to understand intricate contexts. Even without altering their parametric knowledge, LLMs can learn from the provided context \cite{incontext}. This ability, known as in-context learning, enables LLMs to perform a wide range of tasks that require nuanced understanding \cite{incontext}. However, in-context learning can also lead to unintended outcomes, as LLMs may misinterpret the context and generate more hallucinations \cite{context-influence}. Unlike context-conflicting hallucinations, this phenomenon is considered a type of fact-conflicting hallucination, but it is induced by the complexity of the context.

\section{OnionEval}
The objective of OnionEval is to evaluate Small LLMs' fact-conflicting hallucinations across different layers of contextual complexity. Each layer of context is wrapped on the previous question. OnionEval includes 19 categories and 515 entities, comprising a total of 3,356 QA questions designed collaboratively by GPT-4 and human annotators.

We use atomic fact illustrations as the simplest task and gradually layer context onto these atomic fact illustrations. The  O\textsc{nion}E\textsc{val}. generation pipeline consists of four steps: Category and Entity Sampling, Atomic Fact Extraction, Hallucination Generation, and Context Creation.

We use binary classification approach to evaluate SLLMs' hallucination tendencies and the impact of context. Hallucination is defined based on the model's response to a given yes/no question. The performance of OnionEval is evaluated using the Context Influence (\textit{CI}) Score, a metric we proposed to evaluate hallucination induced by input context.


\subsection{Category and Entity sampling}
We initially define 19 categories of entities, including "Vehicles," "Plants," etc. Next, we use ChatGPT to generate a list of the 20 most common entities for each category, such as "Banana" and "Apple" in the "Food" category. Subsequently, with the assistance of human annotators for cross-validation, entities that are overly general are excluded from the list. For instance, "Car" in the "Vehicles" category and "Clothes" in the "Clothing Items" category are removed.

\subsection{Atomic fact extraction}
An atomic fact is a term in philosophy and logic used to describe a basic, irreducible fact about the world. In contrast to more complex facts, which might depend on or be composed of multiple components, atomic facts are considered the simplest statements that express a basic element of reality. A sentence only contains atomic fact cannot be broken down into other sentences\cite{min2023factscore}. To obtain atomic fact of an entity, we utlized knowledge graph. An atomic fact in a knowledge graph refers to the most granular piece of information that represents a relationship or property between entities in a dataset. A knowledge graph (KG) is a structured representation of real-world entities and their relationships, typically modeled as a directed graph. It consists of three main components: a set of entities (\(E\)), a set of relations (\(R\)), and a set of triplets (\(T\)). Each triplet is represented as \((h, r, t)\), where \(h \in E\) is the head entity, \(r \in R\) is the relation, and \(t \in E\) is the tail entity. The KG can be formally expressed as \(\mathcal{G} = (E, R, T)\). With the help of Google Knowledge Graph, we constructed multiple atomic facts (adjacent neighbors) from the entities in the previous step\footnote{https://cloud.google.com/enterprise-knowledge-graph/docs/search-api}. For exmaple: entity \textit{``Venice''} was input into API and we extracted information \textit{``Venice is a city in northeastern Italy and the capital of the Veneto region. Venice is built on a group of 126 islands that are separated by expanses of open water and by canals. Portions of the Venice city are linked by 472 bridges.''}



\subsection{Hallucination generation}
Based on the atomic fact sentences obtained, we utilized GPT-4 as a hallucination generator. We prompted GPT-4 to hallucinate based on the atomic fact sentences. The hallucination generation rule for GPT-4 was to create an incorrect description of the given atomic fact, ensuring that the hallucinated statement was also an atomic sentence. Subsequently, we conducted a second round of human annotator cross-validation to exclude unclear hallucinations. For example, the fact \textit{``Segway was invented by Dean Kamen''} was hallucinated by GPT-4 into \textit{``Segway was invented by aliens.''}

\subsection{Context Wrapping}
To effectively evaluate a model's ability to handle fact-conflicting hallucinations, it is essential to consider not only its factual accuracy but also its performance across various contextual scenarios. In our framework, we assess the model's performance layer by layer (Figure 3). Each layer, except the initial "atomic" layer, is fully derived from the preceding layer. To facilitate better comparison of context-influenced hallucinations, the inserted elements remain unchanged in subsequent layers. Moreover, the model's binary discrimination process remains consistent, mirroring its approach during atomic fact evaluation.

In the context-wrapping phase, the relevant entity is incorporated into the context to establish a consistent, entity-related framework. In the first layer of context wrapping, the context includes not only the hallucinated atomic sentence but also the factual atomic sentence, which is inserted as part of the overall context.

In our experiment, we encapsulated the atomic fact within a scenario where both the atomic fact and the hallucination are embedded in a book-reading story (Figure 3). In the second layer of context wrapping, the book-reading story is further situated in a scenario where the reader is comfortably seated in a coffee shop while reading. This layered scenario setup enables researchers to effectively evaluate their model's ability to detect hallucinations within a contextual framework.


\subsection{Metrics Explanation}
We use accuracy as the metric to evaluate overall model performance in detecting fact-conflicting hallucinations. In OnionEval, models can generate incorrect hallucination detections in two ways: wrong answer and unmatched answer.

\textbf{Wrong Answer}: The model provides an incorrect binary classification. For example, it may discriminate a hallucinated sentence as no hallucinated, or vice versa. \\
\textbf{Unmatched Answer}: The model provides an answer that does not directly address the binary classification question. For example, when asked whether "Toronto is the capital of Canada," the model might respond with "Canada is a country," which, although true, does not answer the specific question of whether the statement about Toronto is a hallucination. \\
Accuracy is then computed using the following formula:

\begin{align*}
\text{Accuracy} = 1 - \frac{N_{\text{wrong}} + N_{\text{unmatched}}}{N_{\text{total}}}
\end{align*}


We also proposed Context-influence(\textit{CI}) score, a quantitative metric to evaluate the influence of context on a model's propensity for fact-conflicting hallucination. The \textit{CI score} is defined as the sum of the atomic hallucination rate (\(\rho_h\)), which represent model's hallucination rate across the whole benchmark on atomic facts, and the sum of \(\Delta_1\) and \(\Delta_2\), which are the differences for the first and second layers to compare with \(\rho_h\), divided by \(n\), number of layers. Researchers can easily scale this number by adding more layers to OnionEval. As we are focusing on SLLMs, two layers of context wrapping already bring significance to the result. These components are calculated as follows:


\vskip -0.1in
\begin{align*}
    \rho_h &= 1 - \rho_\text{atomic} \quad \text{(atomic hallucination rate)} \\
    \Delta_1 &= \rho_\text{atomic} - \rho_\text{first-layer} \quad \text{(first layer difference)} \\
    \Delta_2 &= \rho_\text{atomic} - \rho_\text{second-layer} \quad \text{(second layer difference)} \\
    \text{\it CI} &= \rho_h + \frac{\sum_{i=1}^{n} \Delta_i}{n} \quad \text{(composite score, where n = 2)}
\end{align*}

By using Context-influence score, researchers can effectively test their model's hallucination detection capability under layered contexts. In OnionEval, we used two layers because we are targeting SLLMs and their performance drops significantly after the first layer (Explain in section 4).



\begin{figure*}[t]
\centering
\begin{minipage}{0.328\textwidth}
    \centering
    \includegraphics[width=\textwidth]{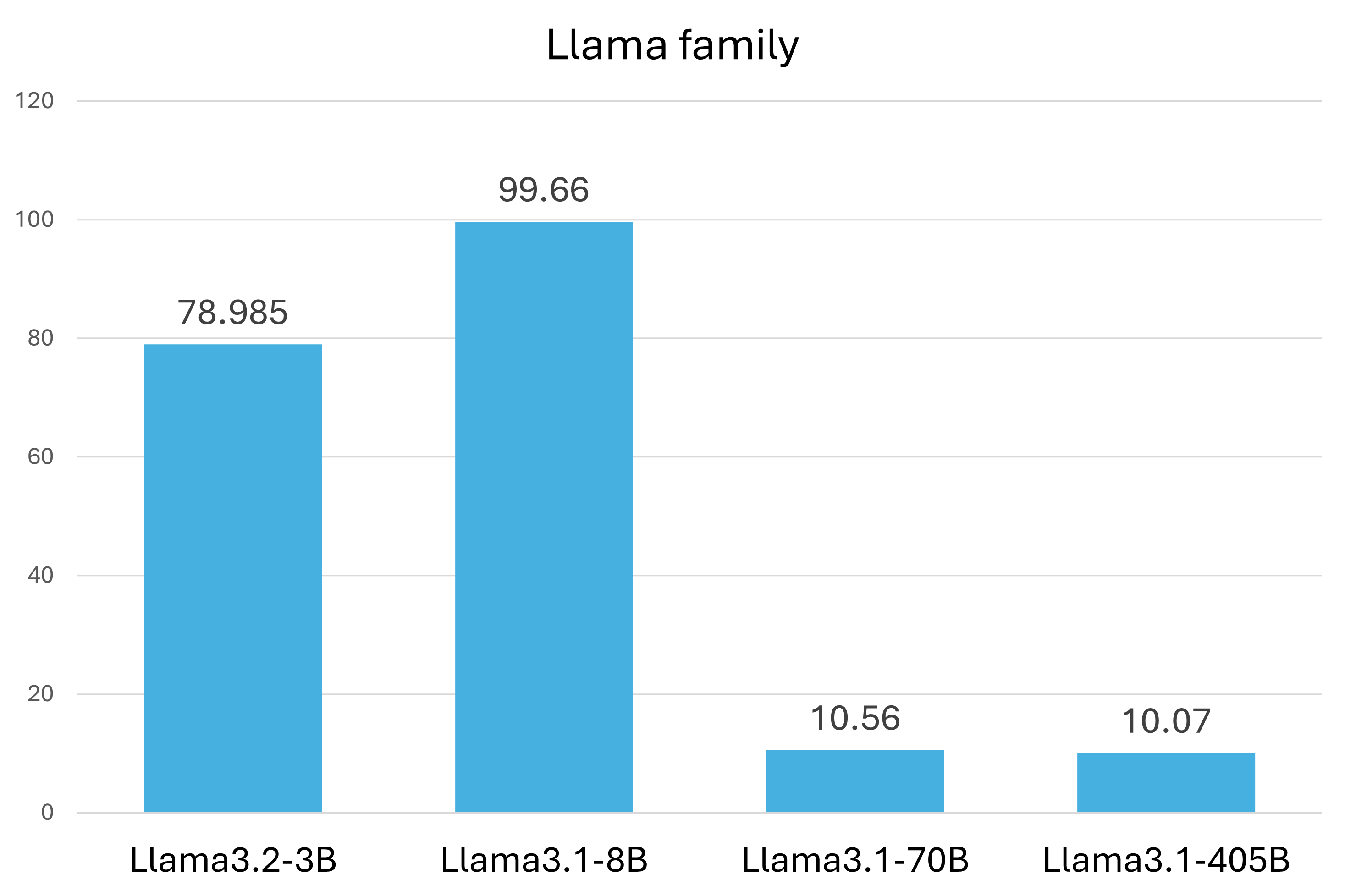}
    \label{fig:llama_family}
\end{minipage}%
\hfill
\begin{minipage}{0.328\textwidth}
    \centering
    \includegraphics[width=\textwidth]{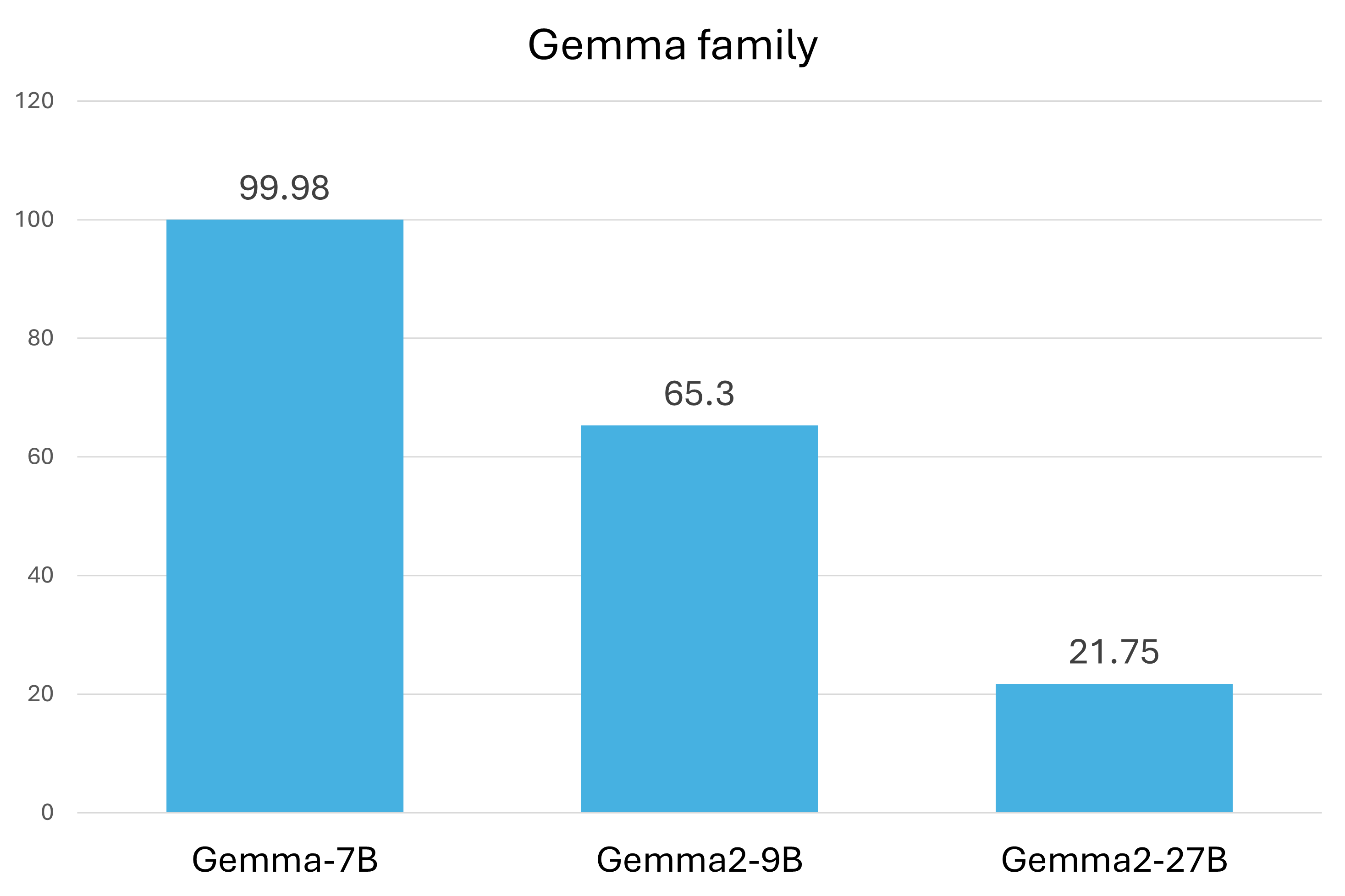}
    \label{fig:gemma_family}
\end{minipage}%
\hfill
\begin{minipage}{0.328\textwidth}
    \centering
    \includegraphics[width=\textwidth]{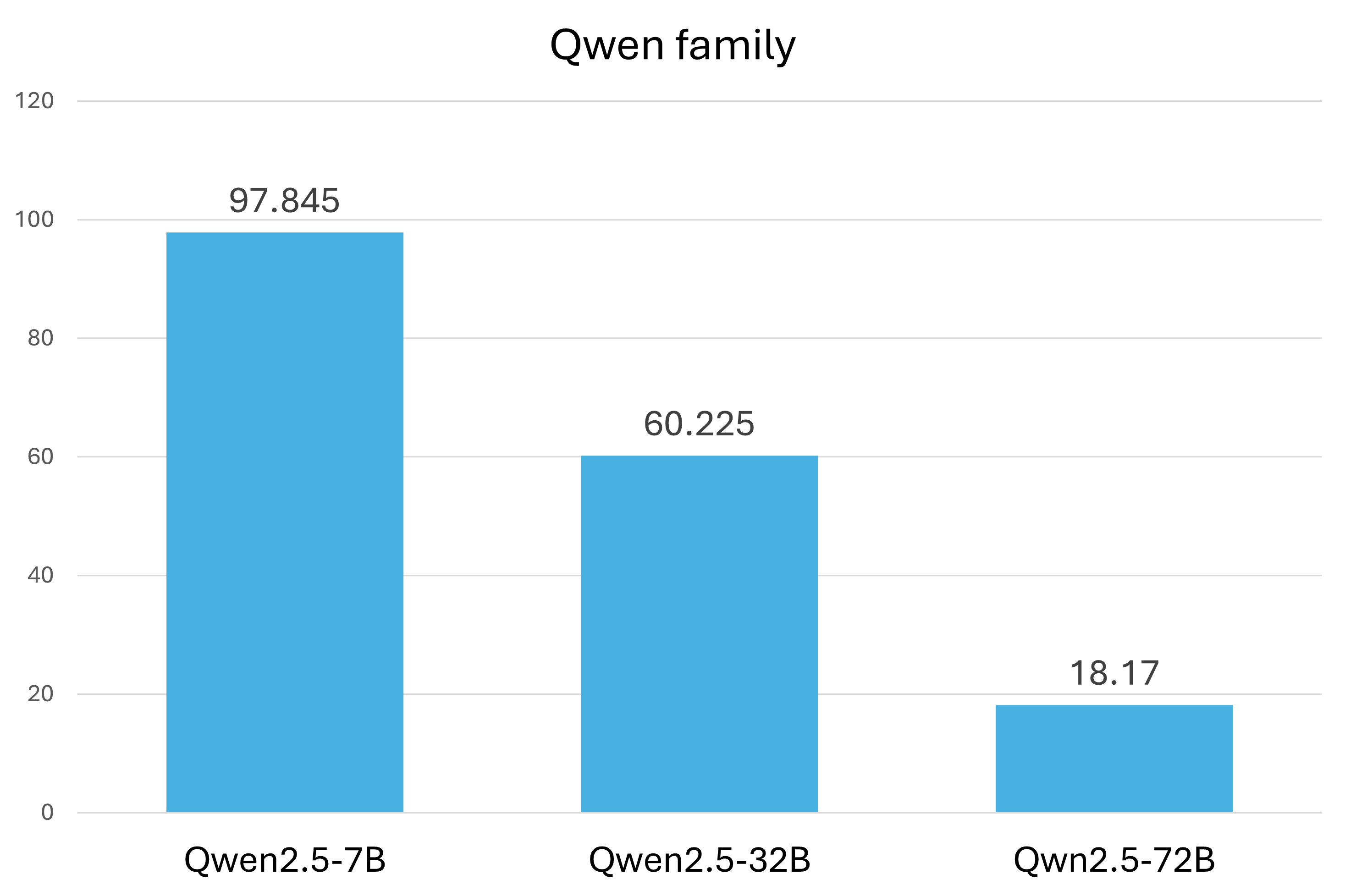}
    \label{fig:qwen_family}
\end{minipage}%

\caption{Context Influence score for models in three families: Llama, Gemma, Qwen. Each subfigure displays the respective Context influence score for different models in the family. From left to right in each figure, the model size increases. However, Context influence score is dropping significantly, which represents larger models having better context-influenced hallucination detection capability.}
\label{fig:universal_caption}
\end{figure*}

\subsection{Benchmark Usage}
Our benchmark can be easily used by researchers. Firstly, based on the atomic sentences which include hallucination, researchers can test their model's capability on atomic fact-conflicting. This step can be used to test as models' ability of fundamental knowledge compression. Secondly, researchers can use the second and third context layer added onto step 1's atomic hallucination. With the context added, researchers can effectively test their models hallucination detection performance on different level of context's complexity. After running the benchmark, researchers can use accuracy and CI score (explained in section 4) to evaluate model's performance on detecting atomic fact information hallucination, and also how small models can be affected by context by CI score. 

\begin{figure}[H]
\vskip 0.1in
\begin{center}
\centerline{\includegraphics[width=\columnwidth]{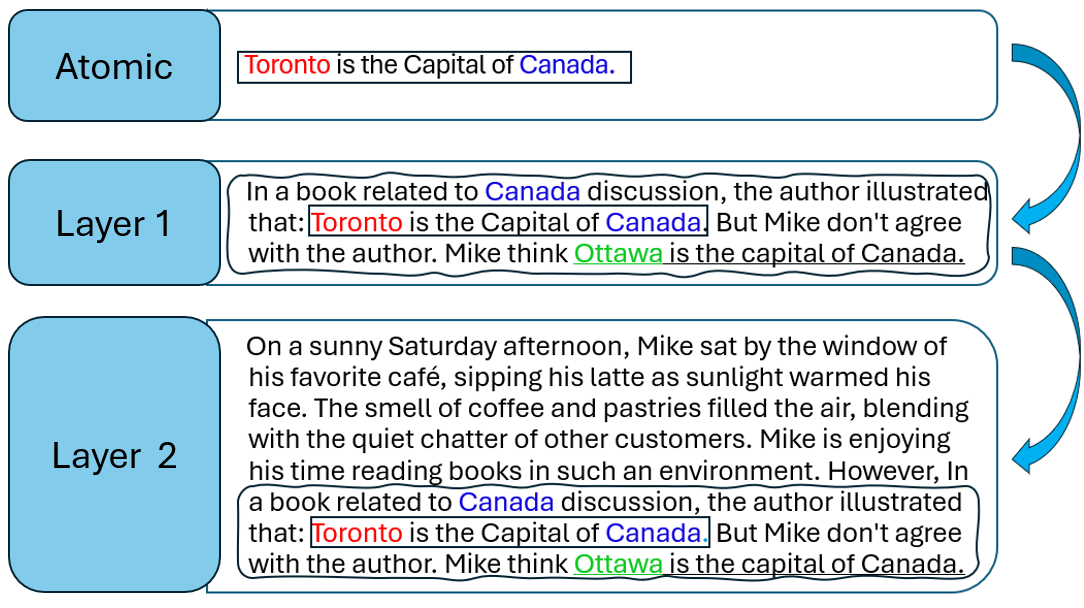}}
\caption{Two levels of context wrapping. The \textcolor{blue}{blue} word is the entity. The \textcolor{red}{red} words represent atomic hallucination words. And, the \textcolor{green}{green} words are atomic facts. Each input is used as part of context of next level's input}
\label{icml-historical}
\end{center}
\vskip -0.5in
\end{figure}

\section{Experiment}
\subsection{Experiment setting}
\textbf{Evaluation Models.}
We evaluated three model families: the Llama 3 (instructed) family, which includes Llama-3 8B, Llama-3.1 8B, Llama-3.1 70B, and Llama-3.2 3B; the Gemma family, comprising Gemma2-27B, Gemma2-9B, and Gemma-7B; the Qwen (instructed) family, which includes Qwen2.5-72B, Qwen2.5-32B, and Qwen2.5-7B. Each of the three families contains at least three models with varying sizes, and all families include a model within the 1B to 10B size range for comparison.

\textbf{Implementation.}
We used the Fireworks API\footnote{https://fireworks.ai/} to evaluate all the models on our framework. The API supports both serverless operation and server-based deployment for models. To eliminate randomness in the responses, we set the temperature parameter of the evaluated models to zero. The output token size was set to 32, as the task required only a binary answer ("yes" or "no"). Additionally, we used the default top-p value of 1.

\subsection{Results}
\begin{table}[h]
\raggedleft
\caption{OnionEval Accuracy, SLLMs' detection accuracy drops significantly after the first and second layer of context wrapping. The \colorbox{lblue}{blue shadow} and \colorbox{lightblue}{lighter blue shadow} in columns represent the performance of SLLMs on atomic fact dataset and context wrapped factuality dataset.}
\label{overall_data}
\vskip 0.15in
\begin{center}
\begin{small}
\begin{sc}
\begin{tabular}{lcccr}
\toprule
Model & Atomic & Layer 1 & Layer 2 \\
\midrule

Llama3.2-3B    & \colorbox{lblue}{90.71} & \colorbox{lightblue}{0.9} & \colorbox{lightblue}{2.6} \\
Llama3.1-8B    & \colorbox{lblue}{98.49} & \colorbox{lightblue}{0.15}  & \colorbox{lightblue}{0.53}   \\
Llama3.1-70B   & 91.46 & 89.08 & 89.80 \\
Llama3.1-405B  & 96.78 & 91.05 & 88.82 \\
\midrule 
Gemma-7B       & \colorbox{lblue}{79.83} & \colorbox{lightblue}{0.03}  & \colorbox{lightblue}{0.012}   \\
Gemma2-9B      & \colorbox{lblue}{94.30} & \colorbox{lightblue}{35.48} & \colorbox{lightblue}{33.93} \\
Gemma2-27B     & 94.10 & 80.74 & 75.77 \\
\midrule 
Qwen2.5-7B     & \colorbox{lblue}{50.48} & \colorbox{lightblue}{1.90}  & \colorbox{lightblue}{2.41} \\
Qwen2.5-32B    & 84.68 & 45.38 & 34.17 \\
Qwen2.5-72B    & 98.27 & 79.05 & 84.62 \\

\bottomrule
\end{tabular}
\end{sc}
\end{small}
\end{center}
\vskip -0.1in
\end{table}

\textbf{Atomic fact hallucination.}
Table~\ref{overall_data} presents the accuracy of LLM evaluations and atomic fact detection across different context levels. The findings reveal that SLLMs perform comparably to their larger counterparts in detecting atomic fact hallucinations. For instance, the Llama3.2-3B model achieves 90.71\% accuracy in atomic hallucination discrimination, closely matching the performance of the larger Llama3.1-70B model, which achieves 91.46\% accuracy.

Moreover, we observed that more recently released models generally outperform earlier versions. For example, the Germma2-9B model achieves a detection accuracy of 94.32\%, surpassing the earlier Gemma-7B model, which has an accuracy of 79.83\%.

We hypothesize that this improvement is due to the use of knowledge distillation in recent models. In the Llama report, developers applied knowledge distillation after model pruning, enabling the distilled model to closely match the performance of its teacher model (e.g., Llama3.1-70B).

\textbf{Context-influenced Hallucination.}

From the result of the experiment, we have found that SLLMs have significantly lower performance in fact-conflicting hallucinations with context. While some SLLMs perform well on atomic fact, but it performs extremely badly on context level 1 and 2. For example, Llama3.1-8B, with 98.49\% accuracy on non-context-wrapped task, only have 0.15\% accuracy on first layer context wrapping and 0.53\% accuracy on second layer context wrapping. However, as a controlled variable for comparison, large models exhibit good performance on both atomic fact task and context-wrapping teak. For example: LLama3.1-70B have three test result very close. This shows that large models have far more better context understanding capabilities than SLLMs. 

Meanwhile, the CI scores for each model family indicate that small language models (SLLMs) are significantly more susceptible to context wrapping effects in hallucination detection tasks. For the LLama family (left panel of Figure 4), the 3B model has a CI score of 78.99, while the 8B model achieves a CI score of 99.66. In contrast, their larger counterparts, the 70B and 405B models, demonstrate relatively better performance, with CI scores of 10.56 and 10.07, respectively.

This trend is also observed in the Gemma and Qwen families. For the Gemma family, the Gemma-7B model (CI score: 99.08) and the Gemma2-9B model (CI score: 65.3) exhibit significantly higher CI scores compared to the larger Gemma2-27B model, which has a CI score of 21.75. A similar pattern emerges in the Qwen family, where the SLLM (Qwen2.5-7B) exhibits a notably higher CI score than the larger 32B and 72B models.

\subsection{Mitigation Strategies}
In this part, we tested three common strategies to mitigate LLM hallucination. We have compared each of the strategies' effectiveness on SLLMs for the first-layer context wrapping. Our findings show that, with correct guidance and prompting, SLLMs can perform as its larger peers. This shows SLLMs is highly capable of doing atomic tasks, but lack of understanding of complex tasks. Model accuracies after mitigation strategy are in table~\ref{aftermitigation}.

\begin{table}[h]
\caption{Llama3.2-3B and Llama3.1-8B performance after mitigation strategies}
\label{aftermitigation}
\vskip 0.15in
\begin{center}
\begin{small}
\begin{sc}
\begin{tabular}{lcccr}
\toprule
Strategy & Llama3.2-3B & Llama3.1-8B \\
\midrule
CoT    & \textbf{74.94} & \textbf{69.45} \\
RAG    & 5.42 & 7.01 \\
One shot    & 2.62 & 15.34 \\
Two shot    & 2.05 & 20.79 \\
Three shot     & 13.7 & 49.88 \\
Four shot      & 10.01 & 61.14 \\
Five shot      & 19.49 & \textbf{66.01} \\
\bottomrule
\end{tabular}
\end{sc}
\end{small}
\end{center}
\vskip -0.1in
\end{table}

\textbf{Chain of thought}

Chain-of-Thought (CoT) is a reasoning framework proposed by Google Brain team\cite{cot}. CoT strategies involve breaking down complex questions or tasks into smaller, more manageable steps and reasoning through them sequentially. Researchers have demonstrated that CoT-like frameworks, such as Chain-of-Verification (CoV), can enhance LLM reasoning and reduce hallucination on the Wikidata dataset \cite{cov}.

In our mitigation experiment, the vanilla Chain-of-Thought approach proved highly effective in improving the accuracy of model hallucination detection. For the Llama3.2-8B model, performance on first-layer hallucination detection increased from 23.92\% to 74.94\%. Similarly, for the Llama3.1-8B model, performance improved dramatically from 0.03\% to 69.45\%.

We hypothesize that SLLMs excel at factual analysis but lack robust internal reasoning capabilities. With the introduction of a structured reasoning process, SLLMs can produce logical reasoning steps that guide them through context-embedded factual analysis effectively. This highlights the potential of reasoning frameworks like CoT to bridge gaps in reasoning and enhance model performance.

\textbf{few-shot prompting}
Few-shot prompting was popularized by researchers in the GPT-3 report \cite{gpt3}. This technique involves providing a model with a small number of examples as part of the input to guide it in performing a specific task. These examples serve as demonstrations, showing the model how to approach a particular type of problem. Common LLMs have demonstrated improved performance in reasoning and task understanding using this approach.

In our experiment, we tested five few-shot prompting setups, ranging from one-shot to five-shot examples. For the Llama3.1-8B model, increasing the number of examples led to a significant improvement in performance. Its accuracy increased from 15.34\% with one-shot prompting to 66.01\% with five-shot prompting. However, the same pattern was not observed for the Llama3.2-3B model. When provided with examples, the 3B model performed worse than the baseline with one-shot prompting. Even with five examples, its performance only matched the baseline.

We conclude that few-shot prompting is not a reasoning-guided strategy but rather a technique for providing hints on how to address a particular type of question. While the 8B model effectively leveraged these hints to detect hallucinations in the test set, the 3B model was unable to benefit significantly, likely due to its smaller capacity for learning from examples.

\textbf{Retrieval Augmented Generation}
Retrieval-Augmented Generation (RAG) combines retrieval and generation to produce accurate, context-aware responses\cite{RAG}. The typical procedure of RAG involves two key components: a retrieval module that gathers relevant information from external knowledge sources and a generative model that uses this information to produce the final output. This approach has been shown to enable large language models (LLMs) to generate explainable, well-reasoned, and less hallucinated results without increasing their parameter size\cite{RAG}.

In our experiment, we used sentence-level retrieval because the hallucination dataset originated from atomic sentences. Therefore, we avoided retrieving very long passages. We implemented RAG for SLLM using the Google Knowledge Graph API. While constructing hallucinated sentences, we also retained the results fetched from the API. These unprocessed sentences served as retrieved information for this experiment.

In our mitigation experiments, models with RAG performed poorly on our layered context hallucination benchmark. For instance, the baseline performance of the Llama3.2-3B model on the first layer of OnionEval was 23.92\%. However, with RAG enhancement, the performance dropped significantly to 5.42\%. Similarly, for the Llama3.1-8B model, the RAG strategy improved performance to only 7, which is still much lower than that achieved with the other two strategies.

Our findings contradict other research claims suggesting that providing external knowledge can mitigate hallucination\cite{li2023halueval}. We conclude that while RAG provides external knowledge, the poor performance of SLLMs on context hallucination benchmarks is not due to a lack of knowledge but rather a lack of reasoning guidance. RAG only introduces additional related knowledge, which, in some cases, misleads the model and results in higher hallucination rates.




\section{Conclusion}

In this study, we introduced OnionEval, a benchmark and framework specifically designed to evaluate fact-conflicting hallucination in Small Large Language Models (SLLMs). Our analysis revealed that while SLLMs demonstrate comparable performance to larger models in detecting hallucinations involving isolated atomic facts, their accuracy declines significantly when these facts are embedded within increasingly complex contextual scenarios. This discrepancy highlights the inherent limitations of SLLMs in context comprehension and reasoning, which larger LLMs handle more effectively.

Through experiments with three prominent model families, Llama, Gemma, and Qwen, we observed that SLLMs struggle with layered contextual tasks due to their limited reasoning capabilities, despite excelling in atomic fact tasks. Moreover, mitigation strategies such as Chain-of-Thought reasoning and few-shot prompting showed promise in improving performance, emphasizing the importance of structured guidance for enhancing SLLMs’ contextual understanding.

Our findings underscore the need for more targeted evaluation methods and model-specific adaptations to bridge the gap between SLLMs and larger counterparts, particularly in tasks that require reasoning across complex or layered contexts. By introducing the Context Influence Score (CI score), we also provide a novel metric for quantifying how contextual complexity affects hallucination detection. Future research should leverage these insights to further explore the limitations and potential of SLLMs, especially in practical deployment scenarios where computational efficiency is paramount.

\section{Limitation}
Although we explored the performance of SLLMs in addressing fact-conflicting hallucinations, research on context-conflicting and input-conflicting benchmarks for SLLMs remains largely unexplored. While our work addressed the gap in hallucination detection within complex contexts, the issue of hallucination in extremely long contexts (e.g., full articles or books) still requires investigation.

Additionally, we observed a trend where newer models generally exhibit better performance. Our benchmark did not account for publication time as a constraint, but future benchmarks could include more detailed evaluations of how training strategies, such as knowledge distillation, influence context-driven hallucination.

\nocite{langley00}

\bibliography{main}
\bibliographystyle{icml2024}


\newpage 
\onecolumn 
\appendix 

\section{Context-Wrapping details}
Context-wrapping details with one example from entity: tulip.

\begin{table*}[h]
    \centering
    \setlength{\tabcolsep}{10pt} 
    \renewcommand{\arraystretch}{1.2} 
    \begin{tabular}{p{0.95\textwidth}} 
        \toprule
        \rowcolor{blue!20}Judge whether the statement I give you is wrong. If it is wrong, reply \texttt{"yes"}. Otherwise, reply \texttt{"no"}. Only reply \texttt{"yes"} or \texttt{"no"} in your answer. \\
        \midrule 
        \rowcolor{green!20}Statement: \textless Tulip flowers are generally black\textgreater.  \\
        \bottomrule
    \end{tabular}
    \caption{Instruction of atomic-fact layer benchmarking. Inside the angle bracket, the text is the atomic fact from OnionEval dataset.}
    \label{tab:comparison}
\end{table*}

\begin{table*}[h]
    \centering
    \setlength{\tabcolsep}{10pt} 
    \renewcommand{\arraystretch}{1.2} 
    \begin{tabular}{p{0.95\textwidth}} 
        \toprule
        \rowcolor{blue!20}I will give you a context about Mike is reading a book. You need to judge if Mike's opinion is correct or not. Think step by step. Answer \texttt{"yes"} if Mike is correct. Answer \texttt{"no"} if Mike is wrong. Only reply \texttt{"yes"} or \texttt{"no"} in your answer.
 \\
        \midrule 
        \rowcolor{green!20}Context: In a book related to [Tulip], the author illustrated that: \textless Tulip flowers are generally black\textgreater.  But, Mike don't agree with the author. Mike think \textless Tulip flowers are generally purple\textgreater.  \\
        \bottomrule
    \end{tabular}
    \caption{Instruction of the first layer context-wrapped benchmarking. Inside the square bracket is the entity. And inside the angle brackets, the texts are hallucination statement and the truth statement from OnionEval dataset.}
    \label{tab:comparison}
\end{table*}

\begin{table*}[h]
    \centering
    \setlength{\tabcolsep}{10pt} 
    \renewcommand{\arraystretch}{1.2} 
    \begin{tabular}{p{0.95\textwidth}} 
        \toprule
        \rowcolor{blue!20}I will give you a context about Mike is reading a book. You need to judge if Mike's opinion is correct or not. Think step by step. Answer \texttt{"yes"} if Mike is correct. Answer \texttt{"no"} if Mike is wrong. Only reply \texttt{"yes"} or \texttt{"no"} in your answer. \\
        \midrule 
        \rowcolor{green!20}Context: On a sunny Saturday afternoon, Mike sat by the window of his favorite café, sipping his latte as sunlight warmed his face. The smell of coffee and pastries filled the air, blending with the quiet chatter of other customers. Mike is enjoying his time reading books in such an environment. However, In a book related to [Tulip], the author illustrated that: \textless Tulip flowers are generally black\textgreater.  But, Mike don't agree with the author. Mike think \textless Tulip flowers are generally purple\textgreater.   \\
        \bottomrule
    \end{tabular}
    \caption{Instruction of the second layer context-wrapping benchmarking. Inside the square bracket is the entity. And inside the angle brackets, the texts are hallucination statement and the truth statement from OnionEval dataset.}
    \label{tab:comparison}
\end{table*}

\begin{table*}[h]
    \centering
    \setlength{\tabcolsep}{10pt} 
    \renewcommand{\arraystretch}{1.2} 
    \begin{tabular}{p{0.95\textwidth}} 
        \rowcolor{blue!20} 
        Judge whether the statement I give you is wrong. If it is wrong, reply \texttt{"yes"}. Otherwise, reply \texttt{"no"}. Only reply \texttt{"yes"} or \texttt{"no"} in your answer. \\
        \rowcolor{green!20} 
        Statement: \textless Tulip flowers are generally black\textgreater. \\
    \end{tabular}
    \caption{Instruction of atomic-fact layer benchmarking. Inside the angle bracket, the text is the atomic fact from OnionEval dataset.}
    \label{tab:comparison}
\end{table*}


\end{document}